\documentclass[transmag]{IEEEtran}
\usepackage{latexsym}
\usepackage{graphicx}
\usepackage{xcolor}
\usepackage{multirow}
\usepackage{amsfonts,amssymb,amsmath}
\usepackage{hyperref}
\def\BibTeX{{\rm B\kern-.05em{\sc i\kern-.025em b}\kern-.08em T\kern-.1667em\lower.7ex\hbox{E}\kern-.125emX}}
\begin{document}

\title{Document Provenance and Authentication through Authorship Classification}

\author{Muhammad Tayyab Zamir\IEEEauthorrefmark{1}, Muhammad Asif Ayub\IEEEauthorrefmark{2}, Jebran Khan, Muhammad Jawad Ikram, \\ Nasir Ahmad, Kashif Ahmad \\
\IEEEauthorblockA{\IEEEauthorrefmark{1} Abasyn University Islamabad Campus} \\
%tayyab.zamir999@gmail.com} \\
\IEEEauthorblockA{\IEEEauthorrefmark{2} DCSE, UET, Peshawar, Pakistan, \\ asifayub836@gmail.com, n.ahmad@uetpeshawar.edu.pk} \\
\IEEEauthorblockA{\IEEEauthorrefmark{3} CSIT, 
Jeddah International College, Saudi Arabia, \\ m.jawad@jicollege.edu.sa} \\
\IEEEauthorblockA{\IEEEauthorrefmark{4} Department of AI, AJOU University, South Korea, \\ jebran@ajou.ac.kr} \\
\IEEEauthorblockA{\IEEEauthorrefmark{5} Department of Computer Science, Munster Technological University, Cork, Ireland\\ kashif.ahmad@mtu.ie} \\
}

\textbf{IEEE Copyright Notice}
Copyright (c) 2023 IEEE
Personal use of this material is permitted. Permission from IEEE must be obtained for all other uses, in
any current or future media, including reprinting/republishing this material for advertising or promotional
purposes, creating new collective works, for resale or redistribution to servers or lists, or reuse of any
copyrighted component of this work in other works.

Accepted to be published in: 2023: IEEE 1st International Conference on Advanced Innovations in Smart Cities (ICAISC 2023), Jeddah, Saudi Arabia, January 23-25, 2023.

%\IEEEmembership{Member, IEEE}
%\thanks{ }
%\thanks{aaaaa).}
%\thanks{bbbbb}
%\thanks{ccccc}

%\thanks{ddddd.}
%\thanks{eeeeee}
%\thanks{fffffff}
%\thanks{gggggg }
%}

\IEEEtitleabstractindextext{\begin{abstract}
Style analysis, which is relatively a less explored topic, enables several interesting applications. For instance, it allows authors to adjust their writing style to produce a more coherent document in collaboration. Similarly, style analysis can also be used for document provenance and authentication as a primary step. In this paper, we propose an ensemble-based text-processing framework for the classification of single and multi-authored documents, which is one of the key tasks in style analysis. The proposed framework incorporates several state-of-the-art text classification algorithms including classical Machine Learning (ML) algorithms, transformers, and deep learning algorithms both individually and in merit-based late fusion. For the merit-based late fusion, we employed several weight optimization and selection methods to assign merit-based weights to the individual text classification algorithms. We also analyze the impact of the characters on the task that are usually excluded in NLP applications during pre-processing by conducting experiments on both clean and un-clean data. The proposed framework is evaluated on a large-scale benchmark dataset, significantly improving performance over the existing solutions. 
\end{abstract}


}

\maketitle
\section{Introduction}
\label{sec:introduction}
Recently the reliability of information spread overshared on online social networks has gained great attention from the community. In this regard, much emphasis is being put on the provenance and authenticity of the information shared on social networks. In the case of written documents, an important aspect of such types of provenance criticism relates to authorship, which involves assessing the authenticity of information by identifying the documents' original author(s) \cite{kestemont2018overview}. Thus, one can argue that the development of computational authorship identification systems, which can assist humans in various tasks in different application domains, such as journalism, law enforcement, content moderation, etc., carries great significance. Style analysis is a primary task in the development of such computational authorship identification systems. Style analysis mainly involves the identification of distinct patterns in writing style by utilizing intrinsic features in a collaboratively authored document. Style change detection (SCD) is one of the key applications of style analysis, which is relevant to intrinsic plagiarism detection (IPD). IPD leverages the writing style consistency of the authors over the length of the document for plagiarism detection independent of any other reference corpus \cite{alvi2022style,bevendorff2022overview}. 

The literature already reports several works on SCD in multi-authored documents. It has also been part of PAN\footnote{https://pan.webis.de/index.html}, which is a benchmark shared task on digital text forensics and stylometry, for several years. The PAN 2021 shared task \cite{zangerle2021overview} on SCD involves three sub-tasks, namely (i) Single vs. Multiple, (ii) Style Change Basic, and (iii) Style Change Real-World. In this work, we target the first task (i.e., single vs. multiple) that involves the classification of single and multi-authored documents. The literature reports several interesting solutions for the task. However, several aspects of the task are still unexplored. For instance, the majority of the solutions rely on individual ML and NLP algorithms. However, based on our previous experience with similar NLP applications \cite{ahmad2022social}, we believe these algorithms can complement each other if jointly utilized in a merit-based fusion. Similarly, the impact of pre-processing step, which generally involves cleaning data by removing unnecessary characters (e.g., stop words, question marks, contractions, etc.,), on the performance of the algorithm in the task is not been explored yet. We believe that the text generally removed in the pre-processing step could be useful in SCD. For example, one author may use stop words, question marks, contractions, etc., more frequently compared to others, which could be useful in SCD.  

To overcome these limitations, we propose a merit-based late fusion-based framework incorporating several state-of-the-art ML and NLP algorithms both individually and jointly using several weight optimization and selection methods for merit-based fusion of the individual algorithms. Moreover, we also analyze the impact of characters that are generally removed during pre-processing in NLP applications on SCD by conducting experiments on both clean and un-clean datasets. The contributions of the work are summarized as follows.
\begin{itemize}
    \item We propose a novel document provenance and authentication framework able to differentiate a document written by single and multiple authors by employing several merit-based late fusion methods.
    \item We evaluate several ML and NLP algorithms including classical and state-of-the-art transformers both individually and jointly in late fusion.
    \item We conducted several experiments to explore different aspects of the task. The experiments are conducted on both cleaned and un-clean data to analyze the impact of characters on style detection that are usually excluded in NLP applications during pre-processing. 
\end{itemize}

%The rest of the paper is organized as follows. Section \ref{sec:related_work} provides an overview of the literature. Section \ref{sec:methodology} describes the proposed methodology. Section \ref{sec:experiments_results} provides an overview of the dataset, experimental setup, conducted experiments, and experimental results. Finally, Section \ref{sec:conclusion} concludes the paper.

\section{Related Study}
\label{sec:related_work}
Stylometry, which involves authors' style analysis, assumes writing style as a quantifiable measure that allows quantifying distinctive features in a given piece of text. Stylometry involves several tasks aiming to analyze several aspects of the text, such as authorship attribution, verification, authorship profiling, stylochronometry, and adversarial stylometry. Authorship attribution is one of the key and widely explored tasks in stylometry \cite{neal2017surveying, stamatatos2009survey}. Author attribution, which aims at the identification of the author of a document, generally relies on SCD techniques.

The literature reports several interesting solutions, exploring different aspects of text analysis, for SCD \cite{lagutina2019survey}. For instance, in one of the initial efforts for the task, Glover et al. \cite{glover1996detecting} employed stylometric features for the identification of author boundaries in documents written by multiple authors. Akiva et al. \cite{akiva2013generic}, on the other hand, proposed a clustering framework for separating a multi-authored document into threads written by individual authors. SCD was also included as a shared task in a benchmark initiative namely PAN style detection 
 for three consecutive years \cite{kestemont2018overview,zangerle2021overview}. Most of the recent works on the topic are based on this shared task. The shared task is composed of multiple sub-tasks including (i) differentiating between single and multi-authored documents and identifying the sequences of paragraphs where the style changes in multi-authored documents considering (i) basic change, and (ii) style
change Real-World. In this paper, we propose a solution for the first task and will provide an overview of the literature on the first task only. 

In the literature, several interesting solutions for the classification of single and multi-authored documents are presented. These solutions explore different aspects of the task. For instance, Zhang et al. \cite{zhangstyle} proposed a BERT and a fully connected neural network-based framework to differentiate between single and multi-authored documents. Strom et al. \cite{strom2021multi} proposed an ensemble-based framework to combine the classification scores obtained through different classifiers trained on two different types of features including BERT embedding and lexical features. The lexical features include character, word, sentence, and contracted and functional word features. Singh et al. \cite{singh2021writing} also relied on lexical features, where a diversified list of features including TF-IDF values of n-grams, POS-Tag tri-grams, and special characters, frequency of function words, and the average number of characters per word, are used. A logistic regression classifier is then trained on the extracted features to classify single and multi-authored documents. Deibel et al. \cite{deibel2021style} used an MLP algorithm trained on a diversified set of textual features including corrected type-token ratio, mean sentence and word length, function word frequency, and linear-write-formula. Nath et al. \cite{nath2021style}, on the other hand, proposed a deep-learning Siamese Neural Network architecture with bi-directional LSTM for the task. 

In this work, in contrast to the existing solutions, we employ several transformers in a merit-based late fusion framework. We also analyze the impact of data cleaning and balancing techniques on the performance of the framework.

%\section{Problem Statement}
%\label{sec:problem_statement}

%Given a text, determine whether the text is written by a single author or by multiple authors. 
 
\section{Methodology}
\label{sec:methodology}
\begin{figure*}[!h]
\centering
\includegraphics[width=0.65\textwidth]{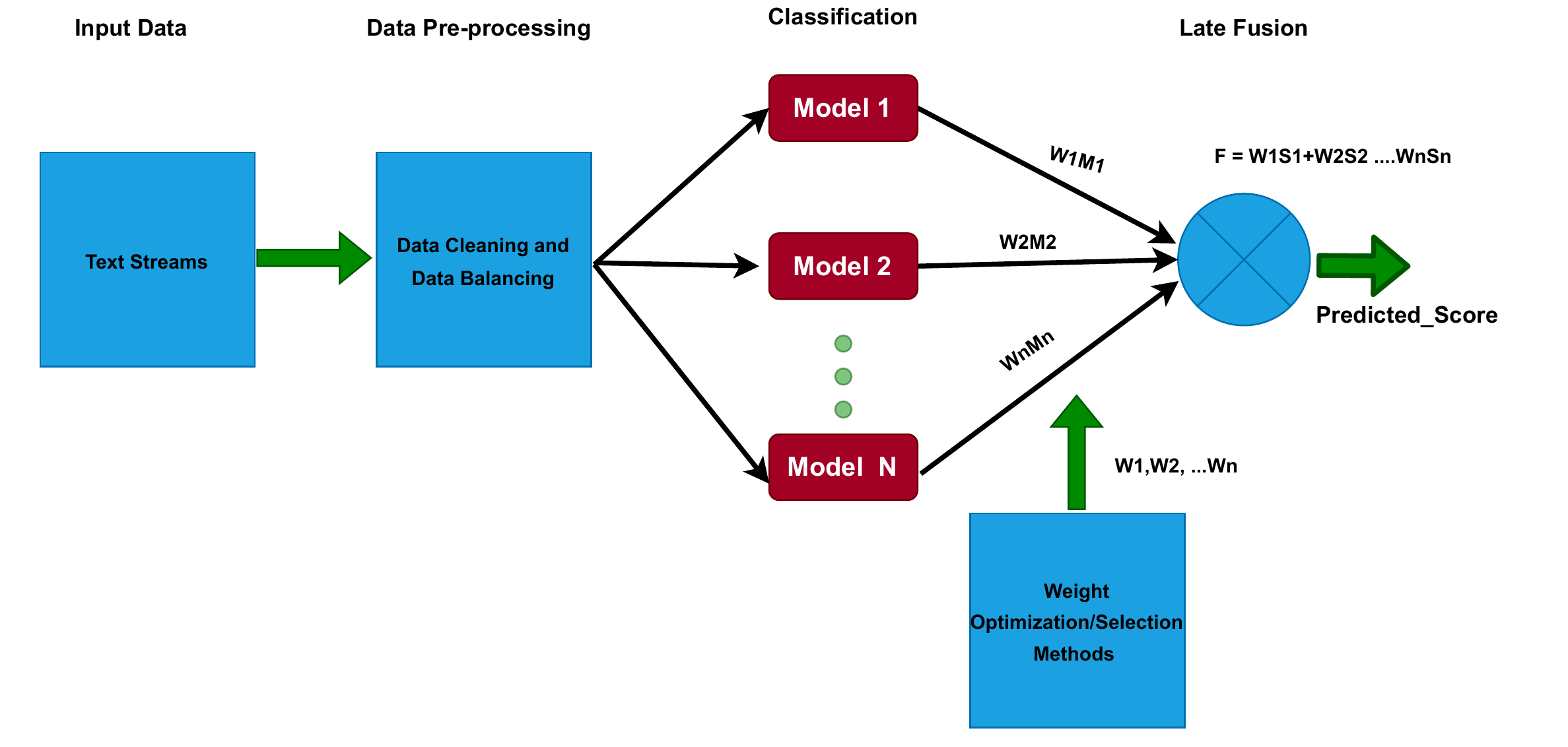}
\caption{A visual depiction of the proposed methodology.} 
	\label{fig:methodology}
\end{figure*}

Figure \ref{fig:methodology} depicts different components of the proposed methodology. The first component is based on several pre-processing techniques to clean and balance the dataset. The pre-processing phase is followed by a classification phase where several algorithms are used. In the final part, we combine the scores of the individual models using several weight optimization and selection methods. The details of these components are provided below.
%%%%%%%%%%%%%%%%%%%%%%%%%%%%%%%%%%%%%%%%%
%%%%%%%%%%%%%%%%%%%%%%%%%%%%%%%%%
\subsection{Pre-processing}
In the pre-processing phase, two different tasks namely (i) data cleaning, and (ii) data balancing are carried out. In the data cleaning task, we removed unnecessary information, such as usernames, URLs, emojis, and stop words. Moreover, we expanded (i.e., decontracted) the contractions using a Python library namely \textit{Contractions\footnote{https://pypi.org/project/contractions/}}.  

In the second part of the pre-processing phase, transpose and Synthetic Minority Oversampling Technique (SMOTE) methods are used to balance the dataset. In the transpose method, new samples are generated by swapping two consecutive/adjacent characters. The SMOTE method, on the other hand, generates newer samples having a closer resemblance to the samples of the minority classes using data augmentation techniques.

\subsection{Feature Extraction and Classification}
In this phase, we employed both classical ML and DL methods. In classical ML methods, we choose several state-of-the-art ML algorithms including Random Forest (RF), Naive Bayes (NB), XGBoost, K-Nearest Neighbour (KNN), Logistic Regression, Decision Trees (DT), and Support Vector Machines (SVMs). We note that all these algorithms are trained on TF-IDF (term frequency and inverse document frequency) features. 

On the other hand, several state-of-the-art models are used in DL methods. These models include BERT (Bidirectional Encoder Representations from Transformers), Roberta (Robustly Optimized BERT), DistilBert, XML-Roberta, XL-Net, and ALBERT. These models are selected based on their proven performances in relevant tasks \cite{said2019natural,ahmad2022social,ahmad2022global}. A brief overview of these models is provided below.

\begin{itemize}
    \item \textbf{BERT}: BERT \cite{devlin2018bert}, which is a transformer-based algorithm, is one of the state-of-the-art NLP algorithms. It is originally pre-trained on a large collection (Wikipedia) of unlabeled text and can be fine-tuned for any NLP task. BERT has several key characteristics that make it a suitable choice for several NLP applications. However, its strength mainly lies in its ability for bidirectional training. This allows it to read the text in both directions simultaneously. This bi-direction training helps in the extraction of contextual information. BERT is available in two different versions having a different number of layers and parameters. In this work, we considered the model having 12 layers and attention heads with a total of 110 million parameters. 
    \item \textbf{DistilBERT}: DistilBERT \cite{sanh2019distilbert} is a smaller, faster, cheaper, and lighter version of the BERT model. The model is trained by distilling the BERT base. The model is able to preserve the 95\% performance of the original BERT model on the GLUE benchmark dataset with a significant (around 40\%) reduction in the number of parameters \cite{sanh2019distilbert}. The reduction in the number of parameters results in a 60\% faster model than the original BERT base model. 
    \item \textbf{RoBERTa}: RoBERTa \cite{liu2019roberta} is a modified form of BERT, where key hyperparameters of BERT are modified. For instance, it is trained with a higher learning rate and larger mini-batch size. More importantly, it removes BERT’s next-sentence pretraining objective resulting in an improved masked language modeling objective compared with the original BERT. Moreover, RoBERTa is trained on a larger collection of data.  
     \item \textbf{XML-RoBERTa}: XLM-RoBERTa is a multilingual version of RoBERTa. The model is trained on a collection of 2.5TB text documents from 100 different languages. The selection of the model for this task is based on its proven performance in similar tasks \cite{ahmad2022social}.
      \item \textbf{ALBERT}: ALBERT \cite{lan2019albert}, which is considered a lighter version of BERT, is built on top of the BERT base model by targeting three aspects. These include  (i) parameter sharing, (ii) embedding factorization,  and (iii) sentence order prediction. For parameter sharing, instead of using different parameters for all of the 12 layers, ALBERT uses the same weights/parameter values for all the layers. This results in a significant reduction in the parameters of ALBERT compared to the original BERT model. ALBERT also uses a reduced size of embeddings. Moreover, in contrast to the original model, ALBERT uses sentence order prediction (SOP) loss based on the coherence of sentences rather than just predicting the topics. 
   % \item \textbf{LSTM}: LSTM is a special type of Recurrent Neural network (RNNs) and is able to learn long-term dependencies. The ability to learn long-term dependencies makes them a suitable choice in a diversified list of applications, such as speech translation. LSTMs have also been widely used in text processing applications \cite{leevy2020short}. Similar to all RNNs, LSTMs also possess chain-like structures, however, in contrast to other RNNs, the repeating module of LSTMs has a different structure by having four neural network layers instead of a single one. 
  %  \item \textbf{biLSTM}: Bidirectional LSTM (biLSTM) is one of the key variants of LSTM. In contrast to standard LSTMs, input flows in both directions in biLSTMs. In other words, we can say biLSTMs are composed of two LSTMs where one takes the input in a forward and the other one takes the input in a backward direction. Such a structure makes it a suitable choice for NLP tasks. 
\end{itemize}

\subsection{Fusion}
In the fusion phase, the scores of the individual models are combined in a late fusion scheme using Equation \ref{eqn:fusion}. In the equation, $S$ is the combined classification score while $s_{n}$ represents the classification score of the nth model. $w_{1},w_{2},w_{3}, ...w_{n} $ represent the weights assigned to the individual models, which are calculated using several weight optimization and selection methods. 
%%%%%%%%%%%%%%%%%%%%
\begin{equation}
\label{eqn:fusion}
S=w_{1}s_{1}+w_{2}s_{2}+w_{3}s_{3}+....+w_{n}s_{n}
\end{equation}
%%%%%%%%%%%%%%%%%%%%%%%%
For weight selection, mainly three merit-based fusion methods namely PSO, NelderMead, and Powell's method are used. In the fusion methods, the weights of the models are optimized/selected on a validation set. In all of the fusion methods, the fitness function is based on an accumulative classification error as shown in Equation \ref{fitness_function}. In the equation, $e$ and $A_{acc}$ represent the classification error and accumulative accuracy, respectively. 
%%%%%%%%%%%%%%%%%%%%%%%%%%%%%%%%%%%%%%%%%%%%%%
\begin{equation}
e = 1-A_{acc}
	\label{fitness_function}
\end{equation}
%%%%%%%%%%%%%%%%%%%%%%%%%%%%%%%%%%%%%%%%%
The accumulative accuracy is computed using Equation \ref{equ:accuracy}, where $x_{1}, x_{2}, x_{3}, ... x_{n} $ are the variables (weight values) to be optimized while $p_{1}, p_{2}, p_{3}, ... p_{n} $ \textit{$p_{n}$} represents the probabilities obtained through the individual models on the validation set. 
%%%%%%%%%%%%%%%%%%%%%%%%%%%%%%%%%%%%%%%%%
\begin{equation}
%\small
%\centering
A_{acc} = x(1)*p_{1}+x(2)*p_{2}+... +x(n)*p_{n} 
\label{equ:accuracy}
\end{equation}
%%%%%%%%%%%%%%%%%%%%%%%%%%%%%%%%%%%%%%%%
The selection of the optimization/selection methods is motivated by their proven performance in similar tasks \cite{ahmad2022social}. The details of these methods are provided below. 

\subsubsection{PSO-based Fusion}
Our first merit-based fusion method is based on PSO \cite{kennedy1995particle}. PSO is heuristic in nature, which means that the solution (weights combination in our application) obtained with PSO is not globally optimal. However, the literature shows that the solution obtained with PSO is usually near the global optimal. The working mechanism of PSO is motivated by a group of fish or a flock of birds moving together, where the individual members of the group help each other by sharing their discoveries with the rest of the group leading to the best hunt for the entire group. In PSO, a similar approach is followed in three different steps. Firstly, an evaluation of each of the candidate solutions is carried out based on fitness criteria. Secondly, the personal best and global best values are updated. Finally, the position and velocity of each particle in the swarm are updated accordingly. %In this work, our fitness function is based on an accumulative classification error computed on a validation set using equation \ref{fitness_function}.

\subsubsection{NelderMead Method-based Fusion}
Our second merit-based fusion method is based on the Nelder Mead algorithm, which can be used for n-dimensional optimization problems where 'n' can be 1 or greater than 1. It is a numerical method and can find the maximum or minimum of an objective function, which is based on the accumulative classification error. %The method is used to find the combination of weights to be assigned to the models that result in minimum classification error. 
Our experiments are based on the implementation of the method provided by a Python open-source library, namely, SciPy\footnote{https://scipy.org/}. 

\subsubsection{Powell's Method-based Fusion}
The third method is based on Powell's method, which is a censored maximum-gradients technique. The method starts with an initial guess (weights combination) and starts moving towards a minimum of a function by finding and calculating a distance in a good direction. One of the main limitations of the method is the assumption of the step size in the direction. The method is simple and has been proven very effective in similar tasks \cite{ahmad2022social}. In this work, our implementation is based on the Python open library SciPy\footnote{https://scipy.org/}. Similar to the other methods, our fitness function for Powell's method is defined in equation \ref{fitness_function}.

%\section{Evaluation Metrics}
%\label{sec:evaluation_metrics}

\section{Experiments and Results}
\label{sec:experiments_results}

\subsection{Dataset}
In this work we used the PAN21 authorship analysis dataset \cite{zangerle_eva_2021_4589145}. The dataset is created for SCD task by collecting randomly chosen threads from the popular StackExchange network of Q\&A sites. The dataset consists of 16,000 text documents drawn from sites related to topics focused on technology. The topical scope of the text is limited to ensure topical coherence in the dataset.

The data is cleaned by removing graphical content, URLs, code snippets, quotations, lists, and very short posts. Moreover, the posts that were edited after their submission were also removed from the dataset to ensure the authors' naturalness. After cleaning the text, it is split into paragraphs, while dropping the paragraphs with less than 100 characters. The original dataset for the task of single and multi-authored documents prediction is imbalanced and it consists of 25\% single-authored and 75\% multi-authored documents. The document length ranges from 1,000 to 10,000 words per document. Moreover, each document has at least 2 and at most 100 paragraphs. The dataset is split into training, validation, and test data with a ratio of 70\% (11,200), 15\% (2400), and 15\% (2400), respectively. The training data contains 8,400 multi-authored, and 2800 single-authored documents. Similarly, the test and validation datasets consists of 18,00 multi-authored and 600 single-authored documents.

\subsection{Experimental Setup}
The objectives of this work are multi-fold, and we want to explore different aspects of the task. On one hand, we want to analyze the role of the characters that are usually eliminated in the pre-processing step, such as stop words, URLs, separate alphanumeric characters, and contractions. We believe some of these words especially the use of contraction can help in the detection of style/author changes in a document. Similarly, we also want to analyze and evaluate the performance of several NLP algorithms including classical ML algorithms, such as RF, DT, SVMs, and transformers, such as BERT, RoBERTa, XLM-RoBERTa, and ALBERT in the task. We also aim to analyze and evaluate the performance of different fusion methods in combining the results of the individual models.    

To achieve these objectives, we conducted the following experiments.

\begin{itemize}
    \item We conducted experiments on both clean and raw data.
    \item We also conducted experiments on the imbalanced and balanced dataset, where three different methods are used for balancing the dataset.
    \item We evaluate the performance of several individual models including classical and DL algorithms. 
    \item We evaluate the performance of several fusion techniques including a naive and three merit-based late fusion schemes.
\end{itemize}

\subsection{Results}
Table \ref{tab:individual_models_results} provides the results of our first experiment where we evaluate the performances of the individual models on both clean and un-clean data in terms of F1-score. Moreover, the models are evaluated on balanced and imbalanced datasets. The performances of the individual models in different experimental setups are described below.

\begin{itemize}
    \item \textbf{Classical Vs. Transformers-based Models}: In total, we used 8 different models including 7 classical ML algorithms trained on TF-IDF features, and five transformers. As expected, better results are obtained with transformers compared to the classical methods on all cleaned, un-clean, balanced, and imbalanced datasets. Overall, the highest score is obtained with XML-RoBERTa with an F1-score of 0.84 on un-clean imbalanced data. However, there is no single winner in all types of datasets. Surprisingly, compared to the transformers, some of the classical ML algorithms, such as RF, obtained better results on the clean dataset. This implies that transformers are more suited for extracting meaningful information from complex data compared to the classical NLP and ML algorithms. 
    \item \textbf{Clean vs. Un-clean Dataset}: Interesting behavior has been observed for the models on clean and unclean datasets. The difference in the performance of the transformers on clean and un-clean datasets is significant while comparable results are obtained with the classical models on both sets of data. The highest F1 score obtained on the clean dataset is 0.79 while it is 0.84 on the un-cleaned dataset. The reduction in the performance of the clean data indicates that the pre-processing step results in the removal of words/characters that are useful for the transformers in the author detection task. We note that during the pre-processing phase, we removed different special characters, such as usernames, URLs, emojis, and stop words. We also expanded contractions. In the future, we aim to explore which of the pre-processing techniques resulted in the removal of useful features by employing explainable NLP techniques.
     \item \textbf{Balanced vs. Un-Balanced Dataset}: The performances of the models also vary on balanced and imbalanced datasets. Overall, better results are observed on the imbalanced dataset. Moreover, we evaluate the performance of two different methods, namely (i) SMOTE and (ii) Transpose, for balancing the datasets. There is no clear winner in the data balancing techniques. In some cases, SMOTE resulted in a better score while in others Transpose method obtained better results.
\end{itemize}

%%%%%%%%%%%Individual Model Results %%%%%%%%%%%%%%%%%
% \usepackage{multirow}
\begin{table*}[]
\caption{Evaluation of the individual models on the clean, raw, balanced, imbalanced dataset.} 
\label{tab:individual_models_results}
\begin{tabular}{|c|ccc|ccc|}
\hline
\multirow{2}{*}{\textbf{Model}} & \multicolumn{3}{c|}{\textbf{Un-clean Data}} & \multicolumn{3}{c|}{\textbf{Clean Data}} \\ \cline{2-7} 
 & \multicolumn{1}{c|}{Imbalanced} & \multicolumn{1}{c|}{Balance (SMOTE)} & Balance (Transpose) & \multicolumn{1}{c|}{Imbalanced} & \multicolumn{1}{c|}{Balance (SMOTE)} & Balance (Transpose) \\ \hline
 %& \multicolumn{1}{c|}{} & \multicolumn{1}{c|}{} &  & \multicolumn{1}{c|}{} & \multicolumn{1}{c|}{} &  \\ \hline
 %& \multicolumn{1}{c|}{} & \multicolumn{1}{c|}{} &  & \multicolumn{1}{c|}{} & \multicolumn{1}{c|}{} &  \\ \hline
  %& \multicolumn{1}{c|}{} & \multicolumn{1}{c|}{} &  & \multicolumn{1}{c|}{} & \multicolumn{1}{c|}{} &  \\ \hline
   %& \multicolumn{1}{c|}{} & \multicolumn{1}{c|}{} &  & \multicolumn{1}{c|}{} & \multicolumn{1}{c|}{} &  \\ \hline
 Random Forest & \multicolumn{1}{c|}{0.62} &	\multicolumn{1}{c|}{0.6} & \multicolumn{1}{c|}{0.61} &	\multicolumn{1}{c|}{0.63} & \multicolumn{1}{c|}{0.49} &	\multicolumn{1}{c|}{0.56} \\ \hline

 Naïve bayes	& \multicolumn{1}{c|}{0.43}	& \multicolumn{1}{c|}{0.45}& \multicolumn{1}{c|}{0.43}	& \multicolumn{1}{c|}{0.43}	& \multicolumn{1}{c|}{0.51}& \multicolumn{1}{c|}{0.48} \\ \hline
XGBoost 	& \multicolumn{1}{c|}{0.62}	& \multicolumn{1}{c|}{0.62}& \multicolumn{1}{c|}{0.62}	& \multicolumn{1}{c|}{0.62}	& \multicolumn{1}{c|}{0.48}	& \multicolumn{1}{c|}{0.36} \\ \hline
KNN	& \multicolumn{1}{c|}{0.49}& \multicolumn{1}{c|}{0.48}	& \multicolumn{1}{c|}{0.49}	& \multicolumn{1}{c|}{0.49}	& \multicolumn{1}{c|}{0.51}	& \multicolumn{1}{c|}{0.48} \\ \hline
SVM	& \multicolumn{1}{c|}{0.57}& \multicolumn{1}{c|}{0.46}& \multicolumn{1}{c|}{0.21}	& \multicolumn{1}{c|}{0.57}	& \multicolumn{1}{c|}{0.45}& \multicolumn{1}{c|}{0.47} \\ \hline
Decision Tree	& \multicolumn{1}{c|}{0.51}& \multicolumn{1}{c|}{0.53}	& \multicolumn{1}{c|}{0.51}	& \multicolumn{1}{c|}{0.52}	& \multicolumn{1}{c|}{0.49}	& \multicolumn{1}{c|}{0.46} \\ \hline
Logestic Regression	& \multicolumn{1}{c|}{0.5}	& \multicolumn{1}{c|}{0.48}	& \multicolumn{1}{c|}{0.21}	& \multicolumn{1}{c|}{0.5}	& \multicolumn{1}{c|}{0.51}	& \multicolumn{1}{c|}{0.46} \\ \hline
BERT	& \multicolumn{1}{c|}{0.79}	& \multicolumn{1}{c|}{0.75}	& \multicolumn{1}{c|}{0.79}	& \multicolumn{1}{c|}{0.79}	& \multicolumn{1}{c|}{0.59}	& \multicolumn{1}{c|}{0.79} \\ \hline
Roberta-Base	& \multicolumn{1}{c|}{0.8}	& \multicolumn{1}{c|}{0.7}	& \multicolumn{1}{c|}{0.43}	& \multicolumn{1}{c|}{0.65}	& \multicolumn{1}{c|}{0.59}	& \multicolumn{1}{c|}{0.62} \\ \hline
Albert	& \multicolumn{1}{c|}{0.79}& \multicolumn{1}{c|}{0.79}	& \multicolumn{1}{c|}{0.53}	& \multicolumn{1}{c|}{0.62}	& \multicolumn{1}{c|}{0.62}	& \multicolumn{1}{c|}{0.53} \\ \hline
Distillbert	& \multicolumn{1}{c|}{0.75}	& \multicolumn{1}{c|}{0.72}	& \multicolumn{1}{c|}{0.2}	& \multicolumn{1}{c|}{0.76}	& \multicolumn{1}{c|}{0.6}	& \multicolumn{1}{c|}{0.2} \\ \hline
XLM-Roberta	& \multicolumn{1}{c|}{0.8481}	& \multicolumn{1}{c|}{0.75}	& \multicolumn{1}{c|}{0.79}	& \multicolumn{1}{c|}{0.66}	& \multicolumn{1}{c|}{0.59}	& \multicolumn{1}{c|}{0.79} \\ \hline
\end{tabular}
\end{table*}
%%%%%%%%%%%%%%%%%%%%%%%%%%%%%%%%%%%%%%%%%%%%%%%%%%%
Table \ref{tab:fusion_results} provides the results of the fusion experiments, where we evaluate the performance of several fusion methods. We note that in the fusion methods, we combine the classification score of the transformers only. These methods include three merit-based and a simple averaging fusion method. In the averaging method, all the models are treated equally while in the other methods weights are optimized/selected for the individual models based on their performances. 

As can be seen in the table, overall, better results are obtained with fusion compared to the individual models on both cleaned and un-clean datasets. The improvement over the best individual model on the cleaned dataset is significant. However, less improvement for the fusion method could be observed on the un-clean dataset. 

Moreover, in fusion methods, better results are obtained by merit-based fusion methods compared to the naive fusion method. This shows the efficacy of merit-based fusion methods and insinuates to consider the performance of the individual models in fusion. As far as the evaluation of the merit-based fusion methods is concerned, no significant variation has been observed. Overall, slightly better results are obtained for Powell method-based fusion compared to the other weight optimization/selection methods.
%%%%%%%%% Fusion Results %%%%%%%%%%%%%%%%%%%%%%%%
\begin{table*}[]
\caption{Evaluation of the fusion methods on the clean, raw, balanced, imbalanced dataset.} 
\label{tab:fusion_results}
\begin{tabular}{|c|ccc|ccc|}
\hline
\multirow{2}{*}{\textbf{Fusion Method}} & \multicolumn{3}{c|}{\textbf{Un-clean Data}} & \multicolumn{3}{c|}{\textbf{Clean Data}} \\ \cline{2-7} 
 & \multicolumn{1}{c|}{Imbalanced} & \multicolumn{1}{c|}{Balance (SMOTE)} & Balance (Transpose) & \multicolumn{1}{c|}{Imbalanced} & \multicolumn{1}{c|}{Balance (SMOTE)} & Balance (Transpose) \\ \hline
  Simple Averaging	& \multicolumn{1}{c|}{0.8272}	& \multicolumn{1}{c|}{0.7673}	& \multicolumn{1}{c|}{0.4546}	& \multicolumn{1}{c|}{0.6623}	& \multicolumn{1}{c|}{0.6167}& \multicolumn{1}{c|}{0.2074} \\ \hline
%weighted fusion	0.8513	0.78778	0.52381	0.6557	0.6288	0.5291
PSO Method	& \multicolumn{1}{c|}{0.8486}	& \multicolumn{1}{c|}{0.8031}	& \multicolumn{1}{c|}{0.5301}	& \multicolumn{1}{c|}{0.6511}	& \multicolumn{1}{c|}{0.6007}	& \multicolumn{1}{c|}{0.54} \\ \hline
Nelder Nead Method	& \multicolumn{1}{c|}{0.8491}	& \multicolumn{1}{c|}{0.8031}	& \multicolumn{1}{c|}{0.52381}	& \multicolumn{1}{c|}{0.6439}	& \multicolumn{1}{c|}{0.56575}	& \multicolumn{1}{c|}{0.5397} \\ \hline
Powell Method	& \multicolumn{1}{c|}{0.8523}	& \multicolumn{1}{c|}{0.8031}	& \multicolumn{1}{c|}{0.52381}	& \multicolumn{1}{c|}{0.6412}	& \multicolumn{1}{c|}{0.5657}	& \multicolumn{1}{c|}{0.54} \\ \hline
\end{tabular}
\end{table*}
%%%%%%%%%%%%%%%%%%%%%%%%%%%%%%%%%%%%%%%%%%%%%%%%%%%

We also provide a comparison of the proposed method against the existing solutions in Table \ref{tab:comparison}. As can be seen in the table, all of the merit-based fusion methods have significant improvement over the existing literature. Our best performing fusion method (i.e., Powell Method-based fusion) has an improvement of 5.7\% over the highest-scoring solution in the literature. 

%%%%%%%%Comparison Against existing methods %%%%%%%%%%%%
\begin{table}[]
\caption{Comparison against existing solutions.}
\label{tab:comparison}
\begin{tabular}{|c|c|}
\hline
\textbf{Method} & \textbf{F1-Score} \\ \hline
 Zhang et al. \cite{zhangstyle} & 0.753 \\ \hline
 Strom et al. \cite{strom2021multi} & 0.795 \\ \hline
 Singh et al. \cite{singh2021writing} & 0.634 \\ \hline
  Deibel et al. \cite{deibel2021style} & 0.621 \\ \hline
   Nath et al. \cite{nath2021style} & 0.704 \\ \hline
 Nelder Nead Method-based Fusion (This work) & 0.849 \\ \hline
  PSO-based Fusion (This work) & 0.848 \\ \hline
Powell Method-based Fusion (This work) & 0.852 \\ \hline
\end{tabular}
\end{table}
%%%%%%%%%%%%%%%%%%%%%%%%%%%%%%%%%%%%%%%%%%%%%

\section{Conclusions}
\label{sec:conclusion}
In this paper, we proposed a merit-based fusion framework for the classification of single and multi-authored documents by employing several weight selections and optimization methods for the joint use of multiple NLP algorithms. Moreover, we evaluated the performance of several individual algorithms including classical ML and deep learning models. We also experimented with both cleaned and un-clean data to analyze the impact of characters on style detection that are usually excluded in NLP applications during pre-processing. We also evaluated the performance of multiple data balancing techniques. Overall, better results are obtained for the transformers. A significant improvement in the merit-based fusion techniques over the individual models is noticed. We also noticed that the characters that are generally excluded in NLP applications during pre-processing steps play a key role in style detection. In the future, we aim to extend the work by exploring the other missing aspects of the task, such as combining the scores obtained through classifiers trained on lexical features and DL models.

\bibliographystyle{IEEEtran}
\bibliography{bibliography}

%\begin{IEEEbiography}[{\includegraphics[width=1in,height=1.25in,clip,keepaspectratio]{a1.jpeg}}]{Jebran. Khan} received the B.Sc. and M.Sc. degrees in computer systems engineering from the University of Engineering and Technology at Peshawar, Peshawar, Pakistan. He received his Ph.D. degree in electronics and information engineering from Korea Aerospace University, Goyang, South Korea. He is currently pursuing post-doc at department of Artificial Intelligence, AJOU University, South Korea. 

%His research interests include social networks analysis, AI, NLP, and recommendation systems. He authored and coauthored many quality journal papers. He has been involve in winning two research grants from NRF Korea.
%\end{IEEEbiography}
\end{document}